\documentclass{article}

\usepackage[main, final]{neurips_2026}
\usepackage{booktabs, multirow, graphicx, makecell, pifont, amsmath}

\usepackage[utf8]{inputenc} 
\usepackage[T1]{fontenc}    
\usepackage{hyperref}       
\usepackage{url}            
\usepackage{booktabs}       
\usepackage{amsfonts}       
\usepackage{nicefrac}       
\usepackage{microtype}      
\usepackage{xcolor}         

\title{Learning Visual Speech Representations via Cross-Modal Distillation and Joint Face-Lip Modeling}

\author{
    Jing-Xuan Zhang$^1$\thanks{Corresponding author},~  
    Genshun Wan$^2$, Jianqing Gao$^2$, Jia Pan$^2$, \\
    \textbf{Yumei Zhang}$^{1,3}$, \textbf{Lichen Zhang}$^1$\\
    $^1$School of Artificial Intelligence and Computer Science, Shaanxi Normal University \\
   $^2$iFLYTEK Research, iFLYTEK Co., Ltd.\\
   $^3$Key Laboratory of Intelligent Computing and Service Technology for Folk Song, \\
   Ministry of Culture and Tourism \\
    \texttt{\{jxzhanggg, zhanglichen, zym0910\}@snnu.edu.cn}, \\
    \texttt{\{gswan, jqgao, jiapan\}@iflytek.com}
}

\begin{document}

\maketitle

\begin{abstract}
Lipreading remains challenging due to the inherent ambiguity of visual speech cues and their sensitivity to head pose and video resolution. 
In this work, we propose FLARE, a framework for robust visual speech representation learning via cross-modal distillation and joint face-lip modeling. FLARE leverages semantically rich Whisper representations as the acoustic teacher signal, converting them into discrete distillation targets through a random-projection quantizer and pretraining the model with a masked prediction objective.
To capture complementary visual information, FLARE employs a dual-stream architecture that jointly processes lip region-of-interest (ROI) and full-face inputs, where the lip stream captures fine-grained articulatory motion and the face stream encodes broader facial dynamics. 
During pretraining, auxiliary audio input is incorporated to facilitate cross-modal alignment, and modality dropout is applied to encourage robust visual-only representations. 
For downstream lipreading, FLARE serves as a visual encoder coupled with either a lightweight Transformer decoder or an LLM-based decoder.
Evaluated on LRS2, LRS3, and the out-of-domain WildVSR benchmark, FLARE achieves strong lipreading performance, with word error rates of 11.7\%, 13.2\%, and 34.3\%, respectively.
A progressive design study confirms the contribution of each major component. Further analyses demonstrate that joint face-lip modeling improves robustness under degraded video resolution and large head pose variation, while yielding more phoneme-discriminative visual representations. Code and pretrained models are available at \url{https://github.com/jxzhanggg/FLARE}.
\end{abstract}

\section{Introduction}

Lipreading, also known as visual speech recognition (VSR), refers to the recognition of spoken content from lip movement sequences without relying on acoustic signals. It serves as a complement to audio-based  speech recognition, particularly when audio signals are degraded or unavailable. 
It thus finds wide applications in areas such as human–computer interaction, intelligent device control, and assistive communication systems.

Learning visual speech representations remains challenging, as viseme-to-phoneme mappings are inherently ambiguous and visual inputs are sensitive to head pose and video resolution.
Early approaches relied on visual bottleneck feature extraction~\cite{tamura2015audio,noda2014lipreading,sui2015extracting}, but research has since shifted to end-to-end lipreading models trained directly with text supervision, covering both isolated-word~\cite{8461326, martinez2020lipreading, wand2016lipreading, ma2022training} and sentence-level recognition~\cite{afouras2018deep, ma2021end, ma2023auto}. 
More recently, audio-visual self-supervised learning (AV-SSL) has emerged as a promising paradigm for visual speech representation learning \cite{ma2021lira, shiavhubert, shi22_interspeech, haliassos2022jointly, haliassos2024braven, Zhang2023av2vec}. By pretraining on unlabeled paired audio-visual speech with objectives such as masked prediction \cite{shiavhubert, zhu2023vatlm, Zhang2023av2vec} or contrastive learning \cite{zeng2021contrastive}, these methods learn representations both within and across modalities. The resulting pretrained encoders can then be adapted for lipreading via task-specific decoders \cite{haliassos2024unified, ZHANG2025126741}, integrated with large language models \cite{yeo2024visual, cappellazzo2025large, thomas2025vallr}, or used to provide visual features for speech-model reprogramming approaches \cite{djilali2023lip2vec, prajwal24_interspeech}.

However, existing AV-SSL approaches are subject to several limitations. First, most methods~\cite{shiavhubert, hsu2022u, haliassos2022jointly, zhu2023vatlm, Zhang2023av2vec, haliassos2024braven, haliassos2024unified} are pretrained from scratch to jointly learn audio, visual, and audio-visual representations within a unified framework. While this paradigm has proven effective, it does not directly address how supervision should be structured to most effectively strengthen visual representation learning. Second, most prior work relies solely on a cropped lip region-of-interest (ROI), which can become unreliable under large head pose variation, low video resolution, or imperfect cropping. Although recent studies~\cite{zhang2020can, Zhang2022face, ZHANG2025126741} suggest that full-face input can improve lipreading performance, the complementarity between face-level and lip-level cues remains insufficiently exploited and not fully understood.

In this work, we propose FLARE, a framework for visual speech representation learning that leverages cross-modal distillation and joint face-lip modeling. To exploit fine-grained, semantically rich acoustic knowledge, we adopt Whisper~\cite{radford2023robust} as the teacher model to provide deep audio representations, from which discrete target labels are derived via a random-projection quantizer~\cite{chiu2022self}. FLARE is then pretrained with a masked prediction objective, enabling effective knowledge transfer from large-scale audio models to the visual modality. To capture complementary visual cues, FLARE takes both the full face and the lip region of interest (ROI) as input, processing each through dedicated frontend encoders before fusing the two streams via channel-wise concatenation with dropout regularization. During pretraining, auxiliary audio input is incorporated to facilitate cross-modal alignment, while modality dropout is applied at a carefully selected probability to encourage robust visual-only representation learning.
To enhance both local and long-range temporal modeling, we further strengthen the model architecture with a ConvNeXt~\cite{liu2022convnet}-inspired frontend and a modified Conformer~\cite{gulati20_interspeech} backbone. For downstream lipreading, we employ a lightweight Transformer decoder to enable fair comparison with existing baselines, as well as a large language model (LLM) decoder to investigate the scalability of the learned representations.

We train FLARE on multiple public audio-visual speech datasets and evaluate it on LRS2~\cite{afouras2018deep}, LRS3~\cite{afouras2018lrs3}, and the out-of-domain WildVSR~\cite{djilali2024vsr} benchmark. 
Our best model achieves word error rates (WERs) of 11.7\%, 13.2\%, and 34.3\% on the three benchmarks, respectively, demonstrating strong lipreading performance.
Beyond the main results, we conduct a progressive design study to quantify the contribution of each major component. We further analyze the complementarity of the face and lip streams under varying visual quality and head pose conditions, and visualize the learned representations to provide deeper insight into the proposed framework.

\section{Related Works}

\subsection{Audio-Visual Cross-Modal Distillation}

Many studies have explored cross-modal distillation for audio-visual speech data, leveraging the inherent synergy between audio signals and lip movements. Afouras et al.~\cite{afouras2020asr} propose distilling a lipreading model from a pretrained ASR model using the KL-divergence between the teacher and student CTC posterior distributions. LIBS~\cite{zhao2020hearing} enhances lipreading models by distilling multi-scale knowledge from speech recognizers. Rather than optimizing a specific lipreading model, audio-visual self-supervised learning (AV-SSL) aims to learn general speech representations through pretraining. For instance, LiRA~\cite{ma2021lira} uses the PASE+~\cite{ravanelli2020multi} model for the cross-modal self-supervised learning of visual representations. AV-HuBERT~\cite{shiavhubert, hsu2022u}, AV-data2vec~\cite{Lian2023avdata2vec}, and AV2vec~\cite{Zhang2023av2vec} adopt audio-visual derived targets and share the encoder across the audio, visual, and audio-visual modalities. In contrast, RAVEn~\cite{haliassos2022jointly} and BRAVEn~\cite{haliassos2024braven} employ modality-specific encoders and an asymmetric distillation strategy between audio and video. 
The most closely related work is Zhang et al.~\cite{ZHANG2025111432}, which also leverages multiple speech foundation models for cross-modal distillation. 
However, FLARE differs in two key respects. First, we adopt a simpler and more targeted distillation strategy using a single large-scale external model. Second, rather than jointly optimizing representations across all modalities, an objective that can complicate optimization~\cite{Lian2023avdata2vec}, we explicitly focus on strengthening visual speech representations.

\subsection{Visual Input Selection for Lipreading}
Lipreading systems traditionally use a cropped lip region-of-interest (ROI) as the standard visual input. However, recent studies~\cite{zhang2020can, Zhang2022face} suggest that this practice may discard useful extraoral cues. In particular, using the full face can improve lipreading performance by incorporating complementary information such as head pose and broader facial dynamics. Existing efforts~\cite{wang2024enhancing, ZHANG2025126741} to exploit such multi-scale visual information, however, mainly rely on training separate models and combining them at the decision level. In contrast, our work focuses on early-stage joint modeling of face and lip inputs to learn more integrated visual speech representations.

\section{Proposed Method}

\begin{figure}
    \centering
    \includegraphics[width=1\linewidth]{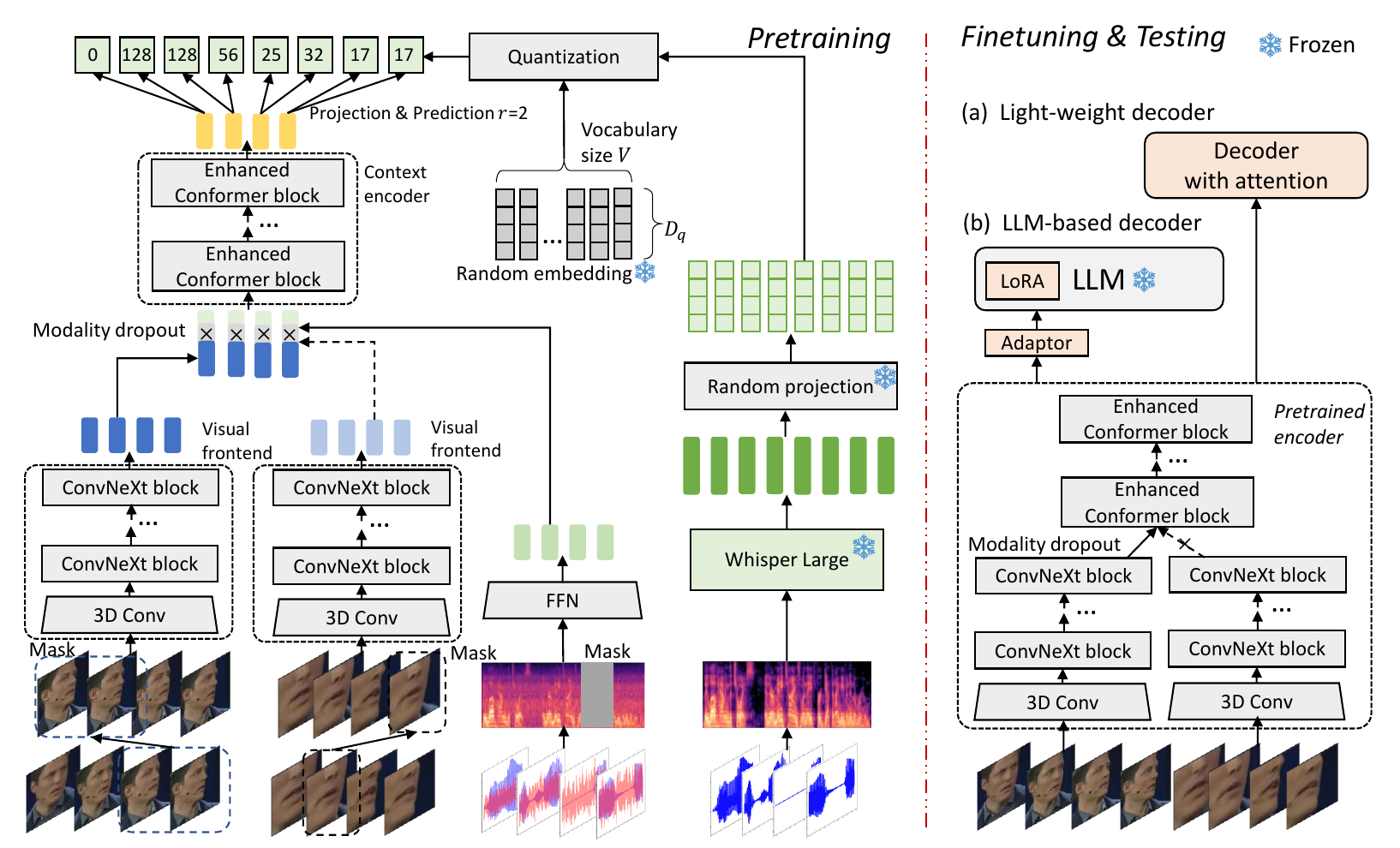}
    \caption{Overview of our proposed method.}
    \label{fig:fig0}
\end{figure}

An overview of FLARE is shown in Figure~\ref{fig:fig0}, and each component is described in the following sections.

\subsection{Model Inputs}

During pretraining, FLARE takes full-face videos, lip region-of-interest (ROI) videos, and auxiliary audio features as inputs.
The face and lip inputs are denoted as $\mathbf{V}^f \in \mathbb{R}^{T \times H \times W}$ and $\mathbf{V}^l \in \mathbb{R}^{T \times H \times W}$, respectively, where $T$, $H$, and $W$ represent the number of frames, frame height, and frame width. 
 For the audio modality, Mel-filterbank features are extracted from the input waveform and denoted as $\mathbf{A} \in \mathbb{R}^{T \times D_a}$.
  Auxiliary audio input is incorporated to facilitate cross-modal alignment, while the input waveform is randomly corrupted with additive noise to prevent over-reliance on acoustic cues.
 
Before being fed into the model, the inputs are corrupted using span-based masking. For visual inputs, following~\cite{shiavhubert}, we randomly replace a span of frames with another span from the same utterance, producing
$\mathbf{V}_m^f = \mathcal{M}_v(\mathbf{V}^f)$ and
$\mathbf{V}_m^l = \mathcal{M}_v(\mathbf{V}^l)$.
Empirically, this strategy performs better than embedding-level feature masking or noise-based input masking.
For audio input, we randomly replace selected frames with a learned embedding, yielding
$\mathbf{A}_m = \mathcal{M}_a(\mathbf{A})$.

\subsection{Model Architecture}

\textbf{Frontend}:
Face and lip videos are encoded by two parallel ConvNeXt~\cite{liu2022convnet}-inspired visual frontends $\text{VF}_f(\cdot)$ and $\text{VF}_l(\cdot)$, respectively, while the audio features are projected by a feed-forward layer:
\begin{equation}
\mathbf{H}^f = \mathrm{VF}_f(\mathbf{V}_m^f) \in \mathbb{R}^{T \times D}, \quad
\mathbf{H}^l = \mathrm{VF}_l(\mathbf{V}_m^l) \in \mathbb{R}^{T \times D}, \quad
\mathbf{H}^a = \mathrm{FFN}(\mathbf{A}_m) \in \mathbb{R}^{T \times D}.
\end{equation}
In each visual frontend, the input video first passes through a 3D convolutional layer to model local spatiotemporal dynamics, followed by four 2D-ConvNeXt blocks. Relative to the widely used 3D-ResNet~\cite{afouras2018deep, shiavhubert}, our design replaces batch normalization with layer normalization, adopts large-kernel depthwise separable convolutions instead of standard small-kernel convolutions, and uses GELU activations~\cite{hendrycks2016gaussian}. 

\textbf{Modality dropout and fusion}:
Before fusion, we apply modality dropout to improve robustness to missing or unreliable modalities.
The probabilities of using audio-only features, video-only features and audio-visual features are $p_a$, $p_v$, $p_{av}$, respectively,  
where $p_a+ p_v + p_{av} = 1$.
For the visual features, the probabilities of using face-only, lip-only, and joint face-lip modeling are $p_f$, $p_l$, and $p_{fl}$, respectively,
where $p_f + p_l + p_{fl} = 1$.
In scenarios where only lip ROI videos are available, such as due to privacy constraints, our framework naturally accommodates missing face inputs by setting the corresponding features to zero during both training and inference.

The remaining audio and visual features are then concatenated along the channel dimension and linearly projected:
\begin{equation}
\label{eq2}
\mathbf{H}^{av} = \mathrm{LN}\big(\mathrm{Concat}(\mathbf{H}^a, \mathbf{H}^l, \mathbf{H}^f) W\big) \in \mathbb{R}^{T \times D},
\end{equation}
where $\mathrm{Concat}(\cdot)$ denotes channel-wise concatenation, $W \in \mathbb{R}^{3D \times D}$ is a trainable projection matrix, and $\mathrm{LN}(\cdot)$ denotes layer normalization.

\textbf{Context encoder}:
The context encoder processes the fused local representations for long-range temporal modeling:
\begin{equation}
\mathbf{H}^o = \mathrm{Enc}(\mathbf{H}^{av}) \in \mathbb{R}^{T \times D}.
\end{equation}
It is critical for lipreading, as sequential context helps reduce the ambiguity of viseme-to-phoneme mapping. We propose an enhanced Conformer as the context encoder. Each block follows the standard Conformer design, consisting of a feed-forward module, a convolution module, a self-attention module, and a second feed-forward module.
To strengthen the encoder, we incorporate several architectural improvements inspired by recent large language models. Specifically, we replace relative positional encoding with rotary positional encoding (RoPE)~\cite{su2024roformer}, adopt grouped-query attention (GQA)~\cite{ainslie2023gqa} for self-attention, use SwiGLU~\cite{shazeer2020glu} in the feed-forward modules, and replace layer normalization with RMSNorm~\cite{zhang2019root}. 
Together, these modifications improve the encoder’s ability to model temporally extended visual patterns while maintaining efficient and stable optimization.

\subsection{Cross-Modal Distillation Loss}

We use the audio-only Whisper model as the teacher to extract deep audio representations from clean speech, denoted by $\mathbf{H}^{\mathcal{T}} \in \mathbb{R}^{rT \times D_{\mathcal{T}}}$, where $r$ is the ratio between the Whisper frame rate and the video frame rate. The teacher representations are globally normalized to zero mean and unit variance, yielding $\bar{\mathbf{H}}^{\mathcal{T}}$.
To obtain discrete supervision targets, we follow a random-projection quantization scheme~\cite{chiu2022self}. Specifically, the normalized teacher representations are first projected into a code space using a random projection matrix $U \in \mathbb{R}^{D_{\mathcal{T}} \times D_q}$, and then assigned to the nearest code in a random codebook $C = [\mathbf{c}_1, \dots, \mathbf{c}_V] \in \mathbb{R}^{D_q \times V}$:
\begin{equation}
y_i = \arg\min_j \left\| \mathrm{Norm}(\mathbf{c}_j) - \mathrm{Norm}\big((\bar{\mathbf{H}}^{\mathcal{T}} U)_i\big) \right\|_2,
\end{equation}
where $\mathrm{Norm}(\cdot)$ denotes $\ell_2$ normalization, $V$ is the codebook size, and $D_q$ is the code dimension. Both $U$ and $C$ are randomly initialized and kept frozen during training. This produces a sequence of discrete target labels $\mathbf{y} = [y_1, y_2, \dots, y_{rT}]$.

To predict these target codes, the model output $\mathbf{H}^o$ is projected to the code vocabulary and reshaped to match the teacher frame rate:
\begin{equation}
P =  \mathrm{Softmax}\big(\mathrm{Reshape}(\mathbf{H}^o W_{\mathrm{proj}}) \big) \in \mathbb{R}^{rT \times V},
\end{equation}
where $W_{\mathrm{proj}} \in \mathbb{R}^{D \times rV}$ is a trainable projection matrix. The $\mathrm{Reshape}(\cdot)$ operation unfolds the channel dimension into $r$ consecutive frames so that the predictions align with the frame rate of the teacher codes.
The model is trained with a cross-entropy loss:
\begin{equation}
L_{\mathrm{pretrain}} = - \frac{1}{rT} \sum_{i=1}^{rT} \sum_{j=1}^{V} \mathbf{y}_{i,j} \log P_{i,j},
\end{equation}
where $\mathbf{y}_{i,:}$ is the one-hot encoding of the target code at frame $i$.
Notably, target codes are predicted for both masked and unmasked regions, as the teacher's deep audio representations provide supervision beyond mere masked reconstruction.
Despite its simplicity, random-projection quantization~\cite{chiu2022self} preserves the geometric structure of the continuous representation space. 
Predicting the quantized codes under masking encourages the model to learn the sequential and structural regularities of the teacher representations, facilitating cross-modal knowledge transfer.
Unlike BEST-RQ~\cite{chiu2022self}, which quantizes Mel-spectrogram features, we quantize deep Whisper representations that encode richer phonetic and linguistic information, providing more informative supervision for cross-modal distillation and yielding stronger visual representations, as validated in Section~\ref{sec:progress}.



\subsection{Lipreading Task Finetuning}

We consider two decoding settings for downstream lipreading: a lightweight Transformer decoder and an LLM-based decoder. The former is used for direct comparison with prior AV-SSL methods, while the latter is adopted to explore the potential of the learned visual representations with a stronger decoder.

For the lightweight setting, the pretrained model serves as the encoder and is paired with a randomly initialized Transformer decoder. We first train the decoder with the encoder frozen, and then jointly finetune both modules. Following most prior AV-SSL work, no auxiliary objective such as CTC loss is used.
For the LLM-based setting, we connect the encoder to the LLM through a lightweight adaptor and adopt low-rank adaptation (LoRA)~\cite{hu2022lora} for parameter-efficient tuning. The encoder is initialized from the model finetuned with the lightweight decoder. A multistage finetuning strategy is used for progressively aligning the encoder outputs with the LLM embedding space. Details are provided in the Appendix~\ref{app:ft}.

\section{Experiments and Results}

\subsection{Experimental Setup}

For pretraining, we use a large-scale audio-visual corpus including LRS2~\cite{afouras2018deep}, LRS3~\cite{afouras2018lrs3}, VoxCeleb2-English~\cite{shiavhubert}, AVSpeech-English~\cite{10.1145/3197517.3201357}, and AVYT-talking~\cite{nguyen25b_interspeech}, totaling 4966 hours after preprocessing. Following prior work, we finetune the pretrained models on both the 30-hour subset and the full set of LRS3, using the LRS3 validation set for hyperparameter tuning. We report results on the test sets of LRS2, LRS3, and the out-of-domain WildVSR benchmark~\cite{djilali2024vsr}. For datasets that only provide lip ROIs, including AVYT-talking and WildVSR, the face features in the modality fusion are set to zero.

We implement two model variants: a \textbf{Base} model with about 117M parameters, using a hidden size of 512 and 12 context encoder blocks, and a \textbf{Large} model with about 342M parameters, using a hidden size of 768 and 18 blocks.
For downstream lipreading, we consider both a lightweight Transformer decoder and an LLM-based decoder (\textbf{FLARE-LLM}). The lightweight setting uses a 6-layer Transformer decoder, whereas the LLM-based setting adopts Qwen-2.5-7B with an adaptor and LoRA for efficient tuning. Additional training details are provided in Appendix~\ref{app:expsetup}.

\begin{table}[t]
\centering
\caption{Lipreading word error rate (WER, \%) on the LRS3 test set, compared with previous AV-SSL baselines under the \emph{low-resource} setting. All methods are pretrained on a combined LRS3 and VoxCeleb2-En corpus.
}
\label{tab:lrs3_compare}
\scalebox{0.85}{
\begin{tabular}{lccccc}
\toprule
\multirow{2}{*}{Method} & \multirow{2}{*}{Encoder size} & \multirow{2}{*}{Ft. criterion} & \multirow{2}{*}{Visual inputs} & \multicolumn{2}{c}{Ft. data} \\
\cmidrule(lr){5-6}
 &  &  &  & 30h & 433h \\
\midrule
\multicolumn{6}{c}{Base size} \\
\midrule
AV-HuBERT~\cite{shiavhubert}            & 103M & CE       & Lip      & 46.1 & 34.8 \\
AV-data2vec~\cite{Lian2023avdata2vec}          & 103M & CE       & Lip      & 37.8 & 32.7 \\
RAVEn~\cite{haliassos2022jointly}               & 97M  & CE+CTC   & Lip      & 40.2 & 33.1 \\
VATLM~\cite{zhu2023vatlm}                & 103M & CE       & Lip      & 42.6 & 34.2 \\
DistillAV~\cite{ZHANG2025111432}            & 103M & CE+KD    & Face     & 36.9 & 31.4 \\
\midrule
\multicolumn{6}{c}{Large size} \\
\midrule
AV-HuBERT~\cite{shiavhubert}            & 325M & CE       & Lip      & 32.5 & 28.6 \\
AV-data2vec ~\cite{Lian2023avdata2vec}          & 325M & CE       & Lip      & 31.6 & 28.5 \\
RAVEn~\cite{haliassos2022jointly}                 & 336M & CE+CTC   & Lip      & 33.1 & 28.2 \\
BRAVEn~\cite{haliassos2024braven}               & 336M & CE+CTC   & Lip      & 30.8 & 26.6 \\
VATLM~\cite{zhu2023vatlm}                 & 325M & CE       & Lip      & 31.6 & 28.4 \\
DistillAV~\cite{ZHANG2025111432}            & 325M & CE+KD    & Face     & 30.2 & 26.2 \\
\midrule
\multicolumn{6}{c}{\emph{Proposed Method}} \\
\midrule
FLARE (\emph{ours})     & 117M & CE       & Face+Lip & \textbf{29.3} & \textbf{24.0} \\
\bottomrule
\end{tabular}
}
\end{table}

\begin{table*}[t]
\caption{
Comparison of lipreading WER (\%) with state-of-the-art baselines on LRS2, LRS3, and the out-of-domain WildVSR dataset under the \emph{high-resource} setting. F+L denotes joint face-lip inputs.
}
\label{tab:vsr_results}
\centering
\scalebox{0.85}{
\begin{tabular}{lccccccccc}
\toprule
Method & \makecell{Encoder\\size} & \makecell{Visual\\input} & \makecell{Data\\size} & LLM & \makecell{LRS2\\WER (\%)} & \makecell{LRS3\\WER (\%)} & \makecell{WildVSR\\WER (\%)} & \makecell{WildVSR\\$\text{Rank}_{wer}\downarrow$} \\
\midrule
\multicolumn{9}{c}{Baselines} \\
\midrule
TM-Seq2seq~\cite{afouras2018deep}   & --   & Lip      & 1391h & \ding{55} & 48.3 & 58.9 & --   & --   \\
ViD 3D~\cite{serdyuk22_interspeech}       & 135M & Lip      & 90kh  & \ding{55} & --   & 17.0   & --   & --   \\
LP-Conformer~\cite{chang2024conformer} & 410M & Lip      & 100kh & \ding{55} & --   & \textbf{12.8} & --   & --   \\
VTP~\cite{prajwal2022sub}          & --   & Lip      & 2676h & \ding{55} & 22.6 & 30.7 & --   & --   \\
Auto-AVSR~\cite{ma2023auto}    & 186M & Lip      & 1902h & \ding{55} & 31.2$^\ddagger$   & 23.5 & 49.3 & 53.2 \\
Auto-AVSR~\cite{ma2023auto}    & 186M & Lip      & 3448h & \ding{55} & 14.6 & 19.1 & 38.6 & 41.8 \\
AV-HuBERT~\cite{shiavhubert}    & 325M & Lip      & 1759h & \ding{55} & 25.5~\cite{zhu2023vatlm}   & 26.9 & 48.7 & 53.0   \\
RAVEn~\cite{haliassos2022jointly}        & 336M & Lip      & 1759h & \ding{55} & 30.2$^\ddagger$   & 23.1 & 46.7 & 49.8 \\
BRAVEn~\cite{haliassos2024braven}        & 336M & Lip      & 2649h  & \ding{55} & 24.4$^\ddagger$  & 20.1  & 38.2$\ddagger$ & 
41.5$^\ddagger$ \\
Llama-AVSR~\cite{cappellazzo2025large}   & 325M & Lip      & 1759h & \ding{51} & 26.9~\cite{cappellazzo2025omni}   & 24.0   & 49.1$^\ddagger$   & 52.5$^\ddagger$   \\
Omni-AVSR~\cite{cappellazzo2025omni}    & 325M & Lip      & 433h  & \ding{51} & 29.5 & 27.1 & 52.9$^\ddagger$   & 55.6$^\ddagger$  \\
UASR-LLM~\cite{zhang2025adapting}     & 325M & Face     & 1759h & \ding{51} & 24.6   & 20.9 & --   & --   \\
\midrule
\multicolumn{9}{c}{\emph{Proposed Method}} \\
\midrule
FLARE       & 117M & F+L & 1759h & \ding{55} & 20.7 & 23.4 & 47.9$^\dagger$  & 52.8 \\
FLARE     & 117M & F+L & 4966h & \ding{55} & 17.3 & 18.5 & 41.1$^\dagger$  & 44.2 \\
FLARE      & 342M & F+L  & 4966h & \ding{55} & 13.6 & 15.9 & 36.9$^\dagger$  & 39.5 \\
FLARE-LLM   & 117M & F+L & 4966h & \ding{51} & 13.1 & 14.4 & 39.7$^\dagger$  & 44.6 \\
FLARE-LLM   & 342M & F+L  & 4966h & \ding{51} & \textbf{11.7} & 13.2 & \textbf{34.3}$^\dagger$  & \textbf{37.6} \\
\bottomrule
\multicolumn{9}{{@{}p{\linewidth}@{}}}{
        $^\dagger$For WildVSR, only lip ROI inputs are used because full-face videos are unavailable. $^\ddagger$These results are obtained by evaluating the publicly released models.
    }
\end{tabular}}
\end{table*}

\subsection{Comparison with Baselines}

We pretrain the Base model on a combined LRS3 and VoxCeleb2-En corpus (1759 hours) and finetune it on the 30-hour and 433-hour LRS3 subsets to evaluate performance in a \emph{low-resource} setting. 
As reported in Table~\ref{tab:lrs3_compare}, our Base model consistently outperforms previous AV-SSL approaches, including those employing encoders of comparable or larger sizes. Notably, it achieves this performance without relying on an external language model, auxiliary CTC supervision, or joint CTC-attention decoding, further demonstrating its effectiveness in visual representation learning.

We further compare our approach with baselines trained on \emph{high-resource} labeled corpora. For our method, the same amount of data (either 1759 or 4966 hours) is used for both pretraining and finetuning.
To provide a more comprehensive evaluation, we also report results on LRS2 and the out-of-domain WildVSR benchmark. In addition to word error rate (WER), we include $\text{Rank}_{wer}$ \cite{djilali2024vsr}, which jointly reflects the mean and variance of WER across test utterances. As summarized in Table~\ref{tab:vsr_results}, FLARE delivers strong performance on both in-domain benchmarks (LRS2 and LRS3) and generalizes effectively to the out-of-domain WildVSR dataset. For WildVSR, we use lip ROI videos as input, since full-face videos are unavailable. Incorporating an LLM-based decoder further improves lipreading accuracy, allowing our method to surpass most previously reported results. Notably, FLARE achieves a WER of 13.2\% on LRS3, approaching the previous best result of 12.8\% reported by \cite{chang2024conformer}, despite being trained on dramatically less data than that system, which uses 100k hours of training data.

\subsection{Progressive Design Study}
\label{sec:progress}

\begin{table}[htbp]
\centering
\caption{Lipreading WER (\%) results of the progressive design study. 
Selected configurations are incorporated into subsequent stages.
All models are pretrained for 200k updates to reduce experimental time.}
\label{tab:pds}
\scalebox{0.85}{
\begin{tabular}{llccc}
\toprule
\multirow{2}{*}{Stage} & \multicolumn{2}{c}{\multirow{2}{*}{Setting}} & \multicolumn{2}{c}{Ft. data} \\ \cmidrule(lr){4-5}
 & \multicolumn{2}{c}{} & 30h & 433h \\ \midrule
\multirow{7}{*}{\begin{tabular}[l]{@{}l@{}}Cross-modal \\ distillation \end{tabular}} & Teacher & Distillation method & & \\ \cmidrule(lr){2-3}
 & WavLM+Conformer ASR~\cite{ZHANG2025111432} & Reg+KLD~\cite{ZHANG2025111432} & 43.7 & 36.5 \\
 & Whisper & Reg+KLD~\cite{ZHANG2025111432} & 43.5 & \textbf{36.3} \\
 & Whisper & Reg & 44.3 & 37.3 \\
 & AV-HuBERT & K-means~\cite{shiavhubert} & 45.4 & 37.0 \\
 & Whisper & K-means~\cite{shiavhubert} & 44.1 & 36.9 \\
 & Mel fbank & Random Projection & 47.8 & 42.5 \\
 & \textbf{Whisper} & \textbf{Random Projection} & \textbf{43.3} & 36.4 \\ \midrule
\multirow{6}{*}{\begin{tabular}[l]{@{}l@{}}Modality \\ dropout \end{tabular}} & \multicolumn{2}{c}{$p_a$ / $p_v$ / $p_{av}$} & & \\ \cmidrule(lr){2-3}
 & \multicolumn{2}{c}{0.25 / 0.25 / 0.5} & 43.3 & 36.4 \\
 & \multicolumn{2}{c}{0.25 / 0.5 / 0.25} & 42.7 & 35.9 \\
 & \multicolumn{2}{c}{0 / 1.0 /0} & 42.3 & 36.2 \\
 & \multicolumn{2}{c}{\textbf{0 / 0.75 / 0.25}} & \textbf{41.0} & \textbf{34.5} \\
 & \multicolumn{2}{c}{0 / 0 / 1.0 $\rightarrow$ 0 / 0.75 / 0.25} & 41.3 & 35.2 \\ \midrule
\multirow{7}{*}{\begin{tabular}[l]{@{}l@{}}Context\\ encoder \end{tabular}} & \multicolumn{2}{c}{Encoder architecture} & & \\ \cmidrule(lr){2-3}
 & \multicolumn{2}{c}{Transformer} & 43.3 & 36.4 \\
 & \multicolumn{2}{c}{Conformer} & 39.2 & 32.3 \\
 & \multicolumn{2}{c}{Enhanced Transformer} & 40.0 & 33.8 \\
 & \multicolumn{2}{c}{Enhanced Conformer} & 38.1 & \textbf{31.8} \\
 & \multicolumn{2}{c}{\textbf{Enhanced Conformer +}} & \multirow{2}{*}{ \textbf{37.0}} & \multirow{2}{*}{32.6} \\
 & \multicolumn{2}{c}{\textbf{Transferred modality dropout}} & &  \\ \midrule
\multirow{4}{*}{Masking} & \multicolumn{2}{c}{Masking strategy} & & \\ \cmidrule(lr){2-3}
 & \multicolumn{2}{c}{Feature mask} & 37.0 & 32.6 \\
 & \multicolumn{2}{c}{Input mask (noise)} & 38.3 & 31.9 \\
 & \multicolumn{2}{c}{\textbf{Input mask (replacement)}} & \textbf{36.3} & \textbf{30.7} \\ \midrule
\multirow{4}{*}{\begin{tabular}[l]{@{}l@{}}Visual inputs \& \\ Frontend\end{tabular}} & Visual frontend & Visual inputs$^\dagger$ & & \\ \cmidrule(lr){2-3}
 & 3D-ResNet & Face & 36.3 & 30.7 \\
 & 3D-ConvNeXt & Face & 36.3 & 30.0 \\
 & \textbf{3D-ConvNeXt} & \textbf{Face+Lip} & \textbf{30.9} & \textbf{25.4} \\ \bottomrule
 \multicolumn{5}{{@{}p{\linewidth}@{}}}{
        $^\dagger$More results on visual inputs
        configuration are reported in Table~\ref{tab:tab4}.
    }
\end{tabular}}
\end{table}

We conduct a progressive design study to disentangle the contributions of key architectural and methodological components. 
At each stage, we progressively refine the system by incorporating selected configurations from preceding stages.
To reduce computational cost, all experiments are conducted using the base model, pretrained on LRS3 and VoxCeleb2-En for 200k updates and then finetuned on the 30h and 433h splits of LRS3.

Table~\ref{tab:pds} summarizes the performance at each stage of this study. We begin by examining cross-modal distillation strategies, comparing Whisper with the WavLM and Conformer ASR ensemble proposed in~\cite{ZHANG2025111432} as teacher models. 
We additionally consider AV-HuBERT~\cite{shiavhubert} as a source of self-supervised audio-visual representations for distillation.
We further investigate several distillation objectives, including regression with KL divergence (Reg+KLD)~\cite{ZHANG2025111432}, standard regression, and target prediction based on either k-means clustering~\cite{shiavhubert} or random projection. The results indicate that different teacher models and distillation objectives lead to broadly comparable performance. However, replacing high-level audio representations with raw Mel filterbanks causes a substantial drop in accuracy, highlighting the importance of semantically richer acoustic supervision. Among the variants, Whisper combined with random-projection quantization achieves the best or similarly strong performance relative to the alternatives.

We then study the effect of modality dropout, parameterized by $p_a/p_v/p_{av}$. The visual-dominant setting (0/0.75/0.25) performs best, surpassing both the balanced setting and the strictly visual-only setting. This finding suggests that although audio provides useful auxiliary guidance during pretraining, stronger emphasis on the visual branch leads to more effective visual speech representations. Introducing a modality-scheduling curriculum from 0/0/1.0 to 0/0.75/0.25 does not yield further gains.

For context modeling, Conformer-based architectures consistently outperform standard Transformers, with our enhanced Conformer achieving the largest improvements. When combined with the transferred modality dropout strategy, VSR performance improves under the 30h finetuning setting but slightly degrades under the 433h setting.
However, after extending pretraining to 400k updates, the transferred modality dropout strategy achieves WERs of 34.0/30.7\%, compared with 35.2/30.6\% for the baseline strategy. These results indicate that the visual-dominant modality dropout strategy remains beneficial with the enhanced contextual encoder under limited finetuning data, while yielding comparable performance with more extensive finetuning data.

We next compare several visual masking strategies, including feature-level masking with a learned mask embedding, input-level masking with Gaussian noise, and input-level masking through intra-utterance span replacement. Among these, span replacement proves to be the most effective by a clear margin. 
Finally, replacing the 3D-ResNet visual frontend with 3D-ConvNeXt yields a modest additional improvement. After extending pretraining to 400k steps, the advantage of 3D-ConvNeXt becomes more pronounced, achieving WERs of 32.2/27.7\%, compared with 33.9/29.5\% for 3D-ResNet. We therefore adopt 3D-ConvNeXt as our visual frontend.

Most notably, jointly modeling face and lip inputs substantially reduces recognition errors, demonstrating that global facial information and fine-grained lip articulation provide highly complementary cues. Additional experimental results 
on visual input configuration are reported in Appendix~\ref{app:visualinput} and Table~\ref{tab:tab4}.
We further analyze this complementarity in the following section.

\begin{figure}[t]
    \centering
    \setlength{\abovecaptionskip}{2pt}
    \setlength{\belowcaptionskip}{-6pt}
    
    \includegraphics[width=0.37\textwidth]{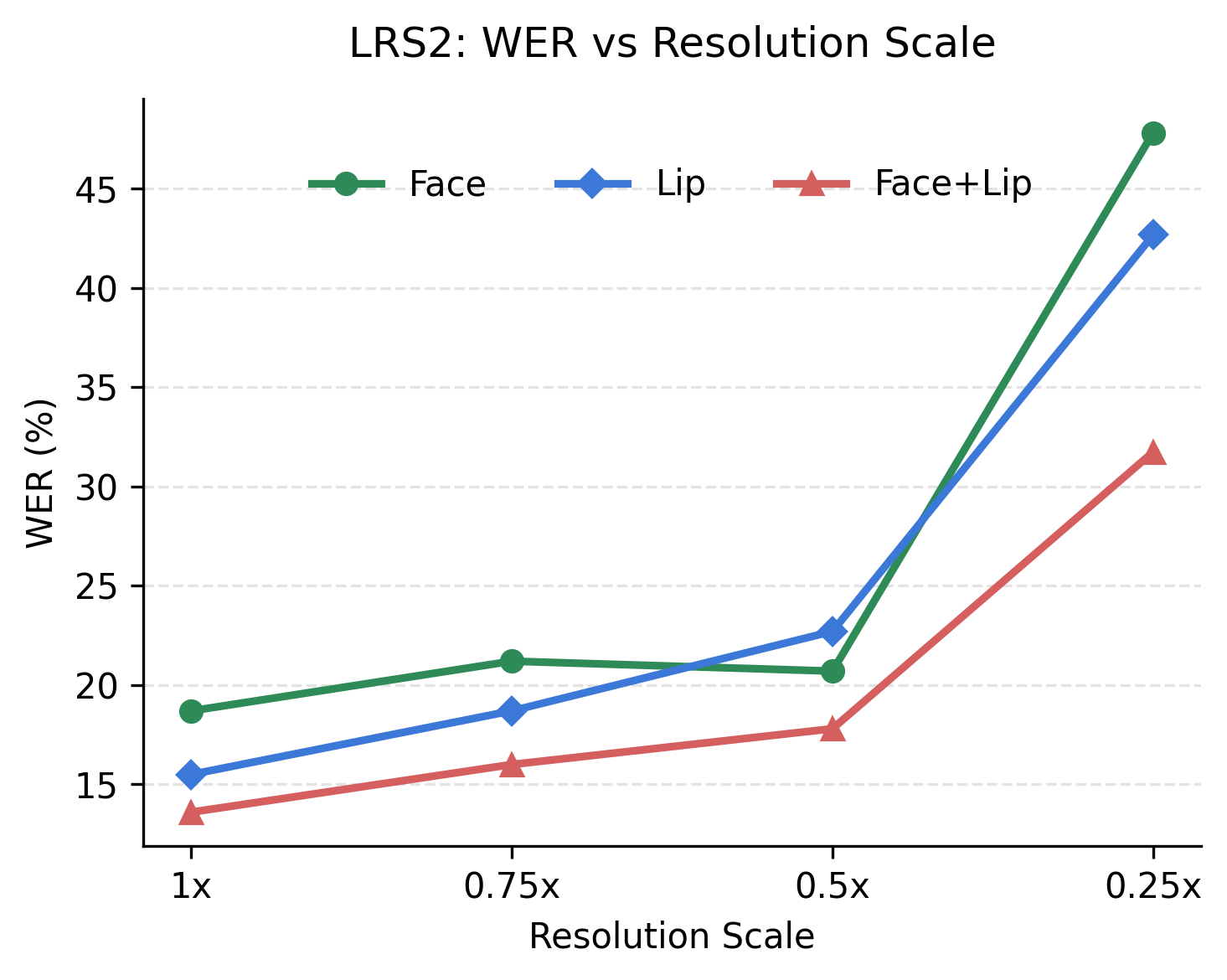}
    \hspace{0.02\textwidth}
    \includegraphics[width=0.37\textwidth]{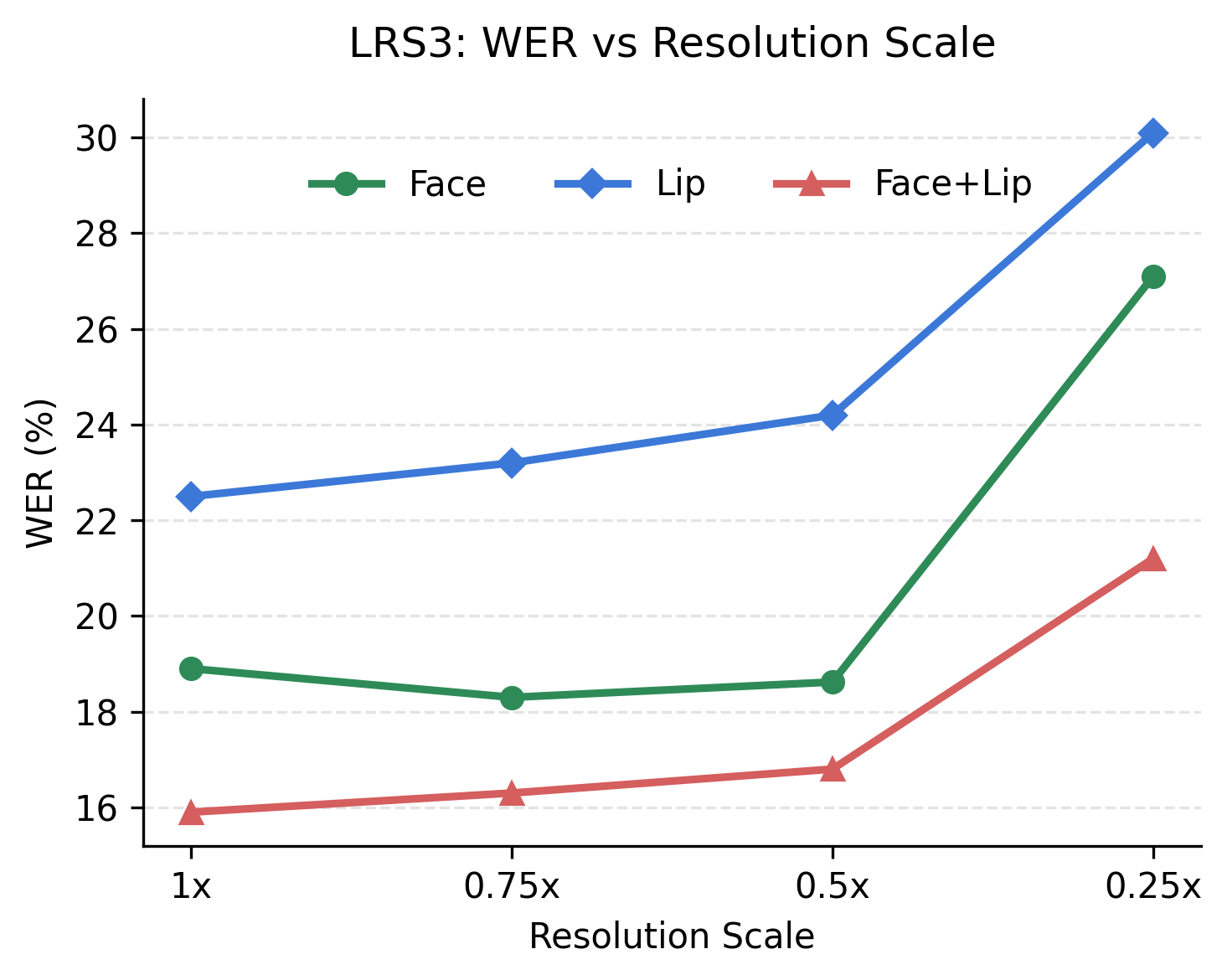}
    
    \caption{Word error rate (WER) comparison across different video resolutions.}
    \label{fig:fig2}
\end{figure}

\begin{figure}[t]
    \centering
     \setlength{\abovecaptionskip}{2pt}
    \setlength{\belowcaptionskip}{-6pt}
    \includegraphics[width=0.37\textwidth]{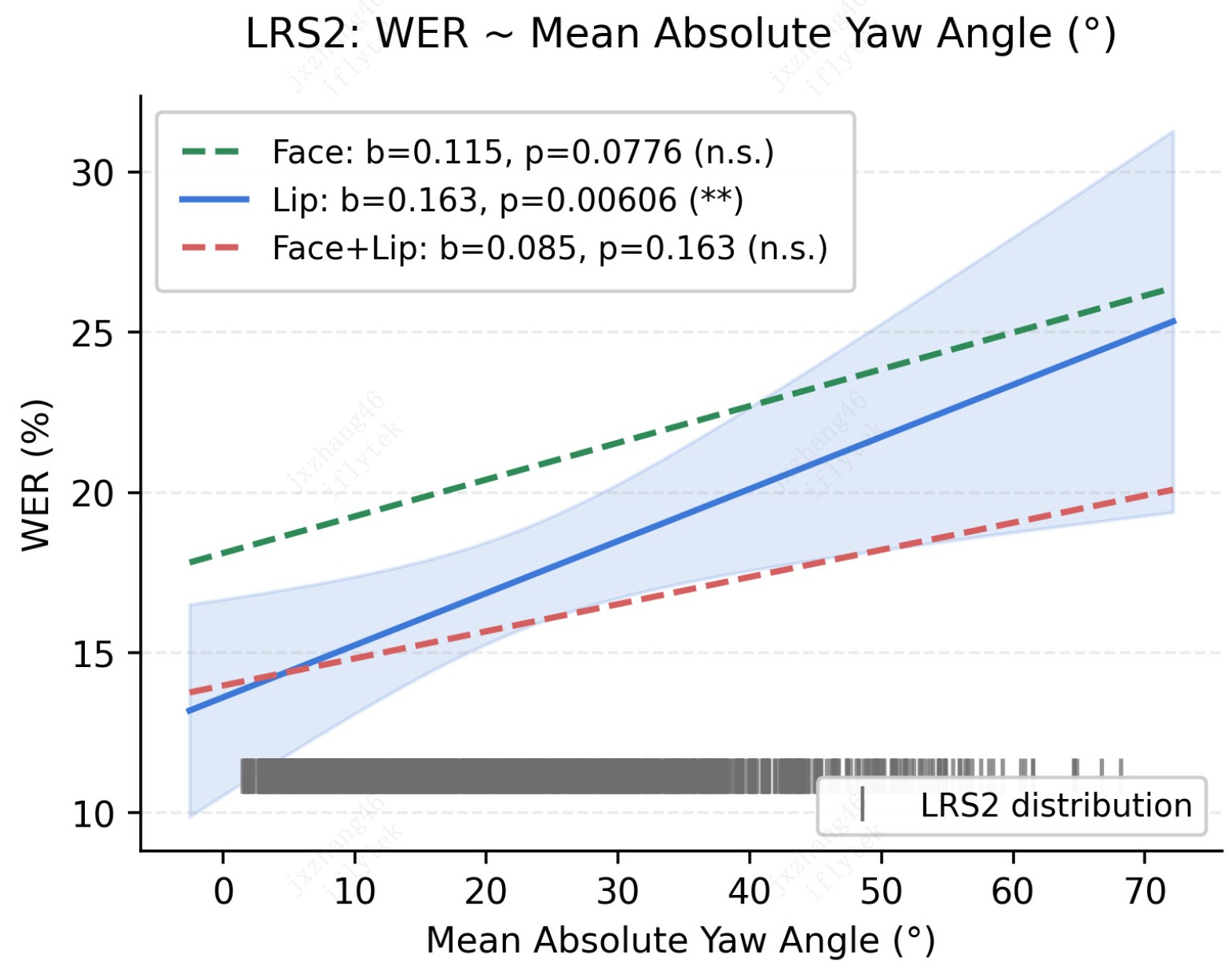}
    \hspace{0.02\textwidth}
    \includegraphics[width=0.37\textwidth]{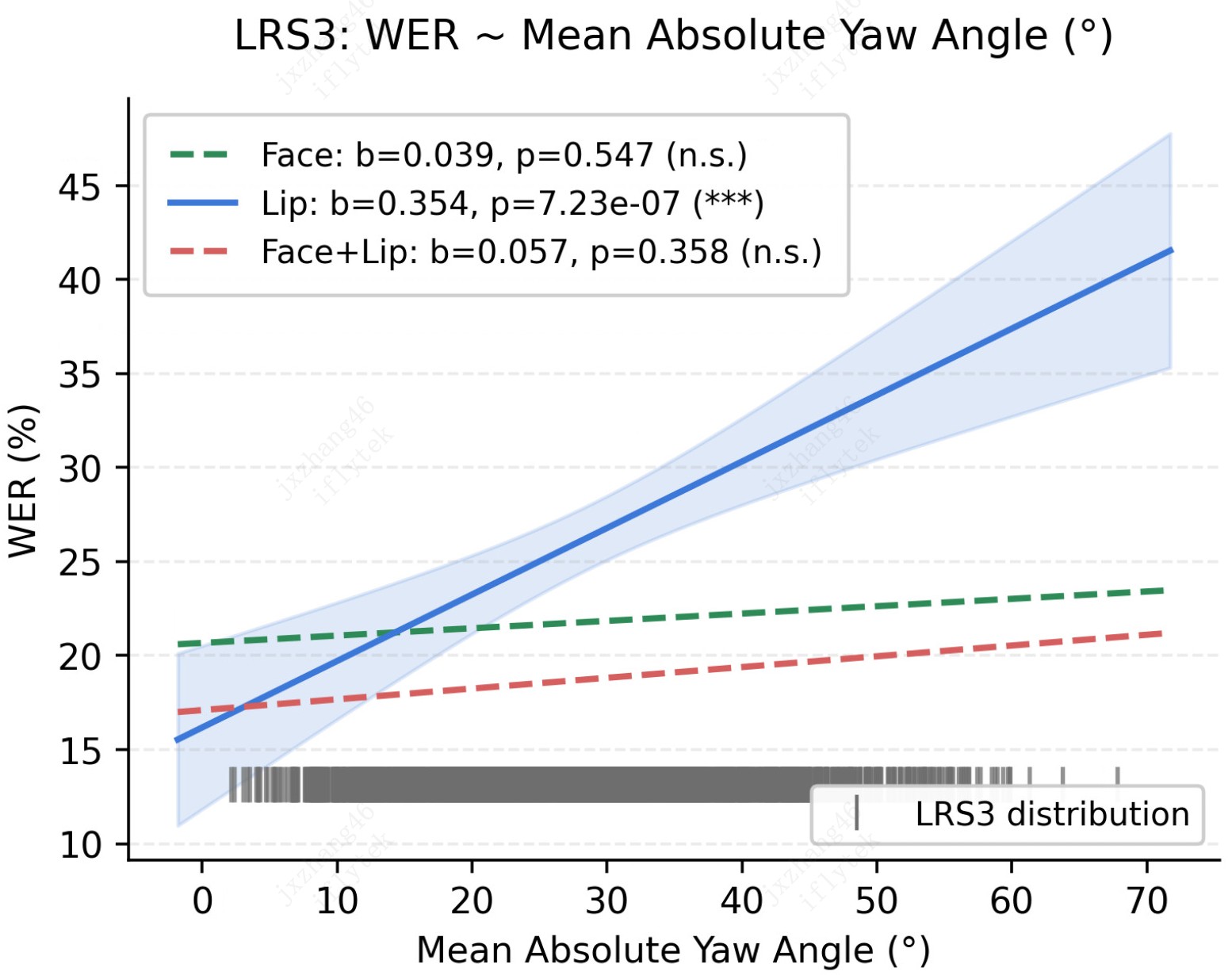}
    \caption{Linear regression analysis of word error rate (WER) as a function of the mean absolute yaw angle.
    }
    \label{fig:fig3}
\end{figure}

\subsection{Robustness to Video Quality and Head Pose}
This section examines the robustness of different visual input configurations under reduced spatial resolution and large head poses. We use the FLARE-Large model  pretrained and finetuned on 4966 hours of data, which supports face-only, lip-only, and joint face-lip inputs.

To evaluate robustness to reduced video resolution, we downsample the original test videos to $0.75\times$, $0.5\times$, and $0.25\times$ of their original spatial resolution. The results in Figure~\ref{fig:fig2} show that face input consistently outperforms lip input on LRS3, whereas lip input generally achieves lower WER on LRS2. This suggests that the relative advantage of face and lip inputs depends strongly on dataset characteristics. Although the broader facial region provides additional dynamic cues beyond the mouth area, its benefit may vary with factors such as pose distribution, visual quality, and background variability. In contrast, joint face-lip input consistently achieves the lowest WER across all settings, indicating that these two input types provide complementary information for robust lipreading.

We next analyze the effect of head pose on recognition accuracy. For each video, we estimate the mean absolute yaw angle and fit a linear regression between yaw and WER, as shown in Figure~\ref{fig:fig3}. When using lip-only input, larger head poses are associated with significantly higher WER on both LRS2 and LRS3. By contrast, face-only and joint face-lip inputs do not show a statistically significant association between WER and head pose. Moreover, joint face-lip input consistently outperforms the individual input types across the full range of head poses. These results suggest that incorporating broader facial information improves robustness under large-pose conditions, where lip region observations alone become less reliable.
Finally, we find that the average mean absolute yaw angle is significantly higher in LRS3 than in LRS2 ($28.7^{\circ}$ vs. $22.6^{\circ}$, $p = 3 \times 10^{-33}$). This difference in pose distribution may partly account for why face input is more advantageous on LRS3, while lip input remains comparatively stronger on LRS2.

\begin{figure}[t]
    \centering
    \includegraphics[width=1.0\textwidth]{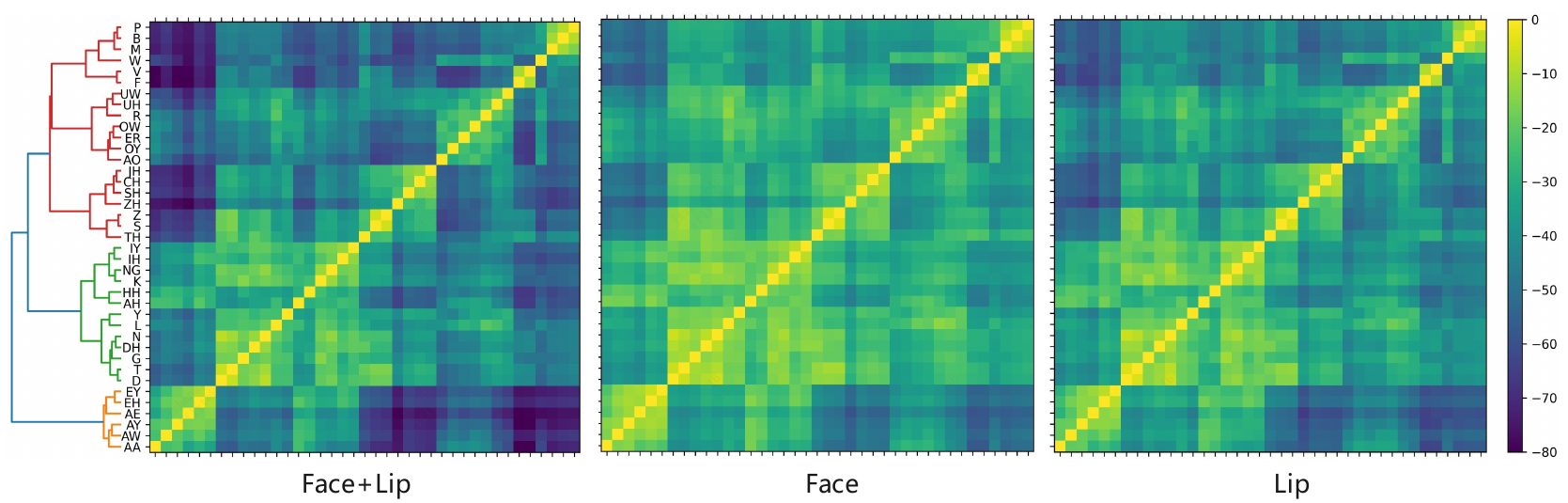}
    \caption{Pairwise negative-distance matrices of phoneme representations for joint face-lip, face-only, and lip-only inputs. Brighter colors indicate smaller distances.}
    \label{fig:fig6}
\end{figure}

\begin{figure}[t]
    \centering
    \includegraphics[width=\textwidth]{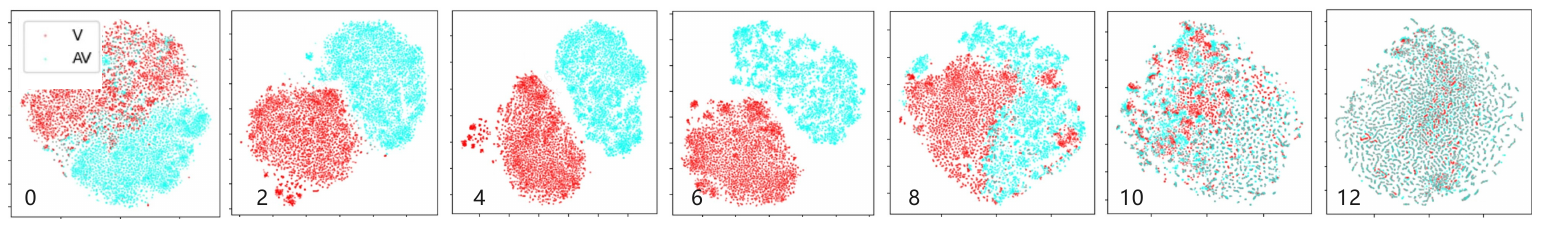}
    \caption{t-SNE visualization of visual and audio-visual representations across context encoder blocks. ``V'' and ``AV'' denote visual-only and audio-visual representations, respectively. The numbers indicate the context encoder depth (0–12).
    }
    \label{fig:fig7}
\end{figure}

\subsection{Visualization of Representations}

We use the FLARE-Base model pretrained on 4966 hours of data in this section.
We first compare the visual representations obtained from face, lip, and joint face-lip inputs on the LRS3 test set. 
The representations are extracted before the context encoder. To analyze their phoneme-level discriminability, we first apply forced alignment to the audio and phoneme transcripts to obtain frame-level phoneme labels. We then compute pairwise phoneme distances in the representation space and visualize the resulting phoneme similarity structure using hierarchical clustering, as shown in Figure~\ref{fig:fig6}. The results indicate that joint face-lip input produces the most phoneme-discriminative representations, exhibiting a clearer and more structured separation among phoneme categories than face-only or lip-only input.

We further visualize the visual and audio-visual representations from different context encoder blocks using t-SNE, as shown in Figure~\ref{fig:fig7}. The results show that, as the depth increases, the visual and audio-visual representations become progressively better aligned. This observation suggests that the context encoder effectively promotes cross-modal consistency, which is conducive to learning stronger visual representations.

\section{Conclusion}

We presented FLARE, a visual speech representation learning framework via cross-modal distillation with joint face-lip modeling. By leveraging semantically rich Whisper representations as supervision and employing random-projection quantization with a masked prediction objective, FLARE learns effective and robust visual representations for lipreading. Parallel face and lip frontends with lightweight fusion further enable the model to exploit complementary information from global facial dynamics and fine-grained articulatory motion. Experiments demonstrate that FLARE delivers strong lipreading performance, with progressive design studies confirming the contribution of each major component. These results establish FLARE as a strong foundation for future research on visual speech understanding, audio-visual speech recognition, and broader multimodal speech tasks.

\begin{ack}
This work was supported by the National Natural Science Foundation of
China under Grant No. 62401348, Fundamental Research
Funds for the Central Universities under Grant No. GK202406005, Young Talent Fund of Association for Science and Technology in Shaanxi, China under Grant No. 20250126.
\end{ack}

{\small
\bibliographystyle{unsrt}
\bibliography{ref}
}







\appendix

\section{Lipreading Task Finetuning Details}
\label{app:ft}
We investigate two approaches for adapting the pretrained model to the lipreading task: a lightweight Transformer decoder with cross-attention, and an LLM-based decoder.

\subsection{Lightweight Decoder}
In this setting, the pretrained model serves as the encoder, while the Transformer decoder is randomly initialized. We first freeze the encoder and train only the decoder for the initial $k$ updates, after which the encoder and decoder are jointly optimized. 
This setting is widely used in prior AV-SSL studies and enables direct comparison with previous methods. To better reflect the quality of the pretrained visual representations, we do not use auxiliary objectives such as CTC loss.

\subsection{LLM Decoder}
This setting is designed to further explore the potential of the learned visual representations with a stronger decoder.
We adopt a multi-stage training strategy. First, we use the encoder obtained after joint finetuning with the lightweight Transformer decoder. Second, we connect the encoder to the LLM through a randomly initialized adaptor. At this stage, both the encoder and the LLM are frozen, and only the adaptor is trained to map visual representations to the LLM embedding space. Third, we introduce low-rank adaptation (LoRA) weights into the LLM and jointly optimize the adaptor and LoRA parameters while keeping all other components fixed.

This stage-wise design gradually aligns the pretrained encoder, adaptor, and decoder, allowing the different components to adapt to one another progressively. Empirically, we find that it stabilizes training and yields better final performance than single-stage optimization.

\section{Experimental Setup Details}
\label{app:expsetup}

For pretraining, we use a large-scale audio-visual corpus consisting of LRS2 ($\sim$223 h, available for non-commercial academic research under a BBC Research \& Development data sharing agreement)~\cite{afouras2018deep}, LRS3 ($\sim$436 h, CC BY 4.0)~\cite{afouras2018lrs3}, VoxCeleb2-English ($\sim$1322 h, CC BY-SA 4.0)~\cite{shiavhubert}, AVSpeech-English ($\sim$1549 h, CC BY 4.0)~\cite{10.1145/3197517.3201357}, and AVYT-talking ($\sim$1437 h, lip-only, non-commercial academic research only; no redistribution)~\cite{nguyen25b_interspeech}, resulting in approximately 4966 hours of data after preprocessing. 
The absence of full-face videos is a dataset-level characteristic. 
AVYT-talking provides only cropped lip ROIs and accounts for approximately 29\% of the total pretraining hours, while the remaining data contain both full-face and lip ROI videos.
For the AVSpeech dataset, we exclude videos whose YouTube IDs overlap with those in the LRS3 test set.
We use MUSAN~\cite{snyder2015musan} (CC BY 4.0) for noise augmentation during pretraining.
To obtain text supervision for finetuning, we use Whisper-Large-v2\footnote{\url{https://huggingface.co/openai/whisper-large-v2}} (MIT License) to identify English utterances and generate transcriptions for VoxCeleb2 and AVSpeech. These transcriptions are used as text labels in the finetuning stage. Since AVYT-talking provides only cropped lip ROIs, we set the face representation $\mathbf{H}^f$ in Equation~\ref{eq2} to zero.
For LRS2, LRS3, and 
VoxCeleb2, we directly use the provided face videos. For datasets that contain only raw videos, we first detect and crop the speaker’s face, and then normalize its scale and spatial alignment to match the preprocessing protocol used in LRS3. To obtain lip inputs, we further detect facial landmarks, align each frame to a canonical face via an affine transformation, and crop the central lip region following prior work~\cite{ma2023auto, shiavhubert}. 
Both the full-face and lip ROI streams are resized to $96\times96$ pixels and randomly cropped to $88\times88$ pixels before being fed into their respective visual frontends. Despite having the same input size, the two streams capture different spatial regions.


For the \emph{low-resource} setting, following previous work~\cite{shiavhubert}], we finetune on both the 30-hour subset and the full set of LRS3. For the \emph{high-resource} setting, pretraining and finetuning use the same amount of data (either 1,759 or 4,966 hours).
The LRS3 validation set is used for hyperparameter tuning. We evaluate on the test sets of LRS2 and LRS3, and further report results on WildVSR~\cite{djilali2024vsr} (CC BY-NC-ND 4.0), a recently introduced out-of-domain lipreading benchmark. Since WildVSR is also lip-only, we likewise set $\mathbf{H}^f$ to zero during evaluation.

For the visual frontend, we use a reduced ConvNeXt-Tiny variant with 14M parameters and 24 layers, which is slightly larger than a typical 3D-ResNet backbone (11M parameters and 18 layers).
We implement two model variants, namely \textbf{Base} and \textbf{Large}. The Base model uses a hidden size of 512 and 12  enhanced Conformer blocks, with 8 query heads and 2 key-value heads in grouped-query attention (GQA), resulting in approximately 117M parameters. The Large model uses a hidden size of 768 and 18 enhanced Conformer blocks, with 12 query heads and 2 key-value heads in GQA, resulting in approximately 342M parameters.
During pretraining, the modality sampling probabilities $p_a$, $p_v$, and $p_{av}$ are set to 0, 0.75, and 0.25, respectively, while $p_f$, $p_l$, and $p_{fl}$ are set to 0.25, 0.25, and 0.5. $V$ and $D_q$ are 8192 and 32 respectively. We optimize the model with Adam. The learning rate is linearly warmed up over the first 10\% of training steps to $5 \times 10^{-4}$, kept constant for the main training phase, and exponentially decayed over the final 10\%. We train for 600k updates on the smaller LRS3+VoxCeleb2 subset (1759 h), and for 800k updates on the full 4966-hour corpus.

For lightweight finetuning, we use a 6-layer Transformer decoder with hidden size 768, resulting in 58M parameters. Training lasts 180k steps, with the encoder frozen for the first 90k steps. The learning rate is linearly warmed up to $1 \times 10^{-3}$ and then exponentially decayed.
For LLM-based finetuning, we use Qwen-2.5-7B\footnote{\url{https://huggingface.co/Qwen/Qwen2.5-7B}} (Apache 2.0) with a two-layer FFN adaptor and rank-16 LoRA, resulting in 51M trainable parameters. The peak learning rate is set to $1 \times 10^{-3}$ during adaptor training and reduced to $2 \times 10^{-4}$ when LoRA is introduced.
The second stage and the third stage finetuning last for 10k updates and 60k updates, respectively. 
Training is conducted on 8 NVIDIA Tesla A100 GPUs. Pretraining requires approximately 3 weeks for the large model and 2 weeks for the base model. Finetuning with a lightweight Transformer decoder takes about 5 and 3 days for the large and base models, respectively. Further finetuning with the LLM requires around 3 and 2 days, respectively.

\section{Effect of Visual Input Configurations}
\label{app:visualinput}

\begin{table}[t]
\centering
\caption{Lipreading WER (\%) on the LRS3 test set for different visual input configurations in the pretraining, finetuning, and testing phases.}
\label{tab:tab4}
\begin{tabular}{l l l c c}
\toprule
Pretraining & Finetuning & Testing & \multicolumn{2}{c}{Ft. data} \\
\cline{4-5}
         &          &      & 30h & 433h \\
\midrule
Face     & Face     & Face       & 36.3 & 30.0 \\
Face+Lip & Face     & Face       & 33.8 & 29.4 \\
Face+Lip & Lip      & Lip        & 38.3 & 32.6 \\
Face+Lip & Face+Lip & Face       & 34.1 & 29.4 \\
Face+Lip & Face+Lip & Lip        & 38.9 & 32.9 \\
Face+Lip & Face+Lip & Face+Lip   & \textbf{30.9} & \textbf{25.4} \\
\bottomrule
\end{tabular}
\end{table}

Table~\ref{tab:tab4} reports the WER results on the LRS3 test set under different visual input configurations used in pretraining, finetuning, and testing. Several observations can be made. First, using joint face-lip inputs during pretraining consistently provides a stronger initialization than using face-only inputs. Second, when combined with our proposed modality dropout, joint face-lip finetuning allows the model to be evaluated with face-only or lip-only inputs at test time while maintaining performance comparable to models finetuned and tested with the corresponding matched inputs. This indicates that joint face-lip training improves input flexibility without sacrificing performance under single-input test conditions. Finally, models finetuned and evaluated with joint face-lip inputs achieve the best overall results, yielding 30.9\% WER on the 30h setting and 25.4\% WER on the 433h setting.

\section{Computational Cost}
\label{app:cost}

\begin{table}[t]
\centering
\caption{Computational comparison of different model architectures and visual frontend configurations. GFLOPs and RTF are measured using a 6-second utterance on a single NVIDIA Tesla P40 GPU.}
\label{tab:efficiency}
\begin{tabular}{llcccc}
\toprule
Model & Encoder & Visual input & Encoder GFLOPs & Total GFLOPs & RTF $\downarrow$ \\
\midrule
\multirow{2}{*}{DistillAV~\cite{ZHANG2025111432}}
    & Base  & Face & 60.6 & 63.2 & 0.18 \\
    & Large & Face & 92.1 & 95.1 & 0.21 \\
\midrule
\multirow{4}{*}{FLARE}
    & Base  & Face     & 66.7  & 69.0  & 0.19 \\
    & Large & Face     & 99.5  & 102.1 & 0.23 \\
    & Base  & Face+Lip & 119.3 & 121.6 & 0.22 \\
    & Large & Face+Lip & 152.3 & 154.9 & 0.23 \\
\midrule
\multirow{4}{*}{FLARE-LLM}
    & Base  & Face     & 66.7  & 2611.7 & 1.31 \\
    & Large & Face     & 99.5  & 2644.5 & 1.36 \\
    & Base  & Face+Lip & 119.3 & 2664.3 & 1.37 \\
    & Large & Face+Lip & 152.3 & 2697.3 & 1.42 \\
\bottomrule
\end{tabular}
\end{table}

We further compare the computational cost of FLARE and FLARE-LLM with that of DistillAV~\cite{ZHANG2025111432}, which shares the same encoder architecture and computational cost as AV-HuBERT~\cite{shiavhubert}. All measurements are conducted using a 6-second utterance on a single NVIDIA Tesla P40 GPU. The real-time factor (RTF) is defined as the ratio of inference time to input duration, with lower values indicating faster inference. The results are summarized in Table~\ref{tab:efficiency}.

The single-frontend FLARE models exhibit computational costs comparable to those of DistillAV, with only marginal increases in GFLOPs and RTF. 
Incorporating an additional lip frontend substantially increases the computational cost despite a relatively modest increase in model parameters. Nevertheless, this additional computation brings notable improvements in recognition accuracy, as demonstrated in the preceding experiments, while maintaining an RTF below 0.24 for both Base and Large configurations.
For FLARE-LLM, the integration of the LLM substantially increases the overall computational cost, with the LLM accounting for the majority of the computation. Consequently, the relative contribution of the encoder becomes considerably smaller, with overall RTF values ranging from 1.31 to 1.42 across different configurations.
As FLARE-LLM employs a 7B-parameter LLM without token compression, improving its computational efficiency remains an important direction for future research.

\section{Limitations}
\label{app:limit}

This work has several limitations. First, joint face-lip modeling introduces an additional visual branch, which increases model complexity and computational cost relative to conventional lip-only approaches. Second, the current study mainly focuses on lipreading accuracy and robustness, without investigating whether the proposed framework transfers effectively to broader audio-visual speech tasks or whether the learned visual representations capture other relevant information, such as speaker characteristics and additional non-linguistic cues. We leave these directions to future work.

\section{Broader Impacts}
\label{app:impact}

This work studies visual speech representation learning for lipreading. 
The positive impact of this research is to improve speech recognition in acoustically challenging environments, such as noisy public spaces, industrial settings, or scenarios where audio is unavailable or severely corrupted. 
It may also benefit assistive communication technologies for people with hearing or speech impairments, and support more robust multimodal human-computer interaction systems.

At the same time, improved lipreading systems may introduce potential societal risks. 
For example, such technology could be misused for covert surveillance or unauthorized inference of speech content from visual recordings, raising privacy concerns. 
In addition, like other speech and vision systems, lipreading models may perform unevenly across different real-world conditions, such as variations in head pose, visual quality, occlusion, recording devices, or demographic and linguistic characteristics. 
Model errors in practical deployments may therefore lead to incorrect transcriptions, unreliable assistance, or unfair performance disparities across users and environments.


\newpage
\section*{NeurIPS Paper Checklist}

\begin{enumerate}

\item {\bf Claims}
    \item[] Question: Do the main claims made in the abstract and introduction accurately reflect the paper's contributions and scope?
    \item[] Answer: \answerYes{} 
    \item[] Justification: 
    The abstract and introduction reflect the paper’s main contributions, including the FLARE framework, Whisper-based cross-modal distillation, joint face-lip modeling, architectural refinements, and empirical evaluation on LRS2/LRS3/WildVSR. These claims are supported by the method and experiments in Sections 3 and 4.

\item {\bf Limitations}
    \item[] Question: Does the paper discuss the limitations of the work performed by the authors?
    \item[] Answer:  \answerYes{} 
    \item[] Justification: 
We discuss the limitations of our work in Appendix~\ref{app:limit}.

\item {\bf Theory assumptions and proofs}
    \item[] Question: For each theoretical result, does the paper provide the full set of assumptions and a complete (and correct) proof?
    \item[] Answer: \answerNA{}
    \item[] Justification: 
    The paper does not present formal theoretical results, theorems, or proofs.

    \item {\bf Experimental result reproducibility}
    \item[] Question: Does the paper fully disclose all the information needed to reproduce the main experimental results of the paper to the extent that it affects the main claims and/or conclusions of the paper (regardless of whether the code and data are provided or not)?
    \item[] Answer: \answerYes{}
    \item[] Justification: 
    The paper discloses the model architecture, datasets, training pipeline, ablation setup, and many implementation choices in the main text and appendices, which supports reproduction of the main claims. 

\item {\bf Open access to data and code}
    \item[] Question: Does the paper provide open access to the data and code, with sufficient instructions to faithfully reproduce the main experimental results, as described in supplemental material?
    \item[] Answer: \answerYes{}
    \item[] Justification: 
    The codebase has been prepared in a double-blind manner for the review process (\url{https://anonymous.4open.science/r/FLARE-522C}) and will be publicly released upon acceptance.  All datasets used in this work are public audio-visual speech datasets, and their details are provided in Appendix~\ref{app:expsetup}.

\item {\bf Experimental setting/details}
    \item[] Question: Does the paper specify all the training and test details (e.g., data splits, hyperparameters, how they were chosen, type of optimizer) necessary to understand the results?
    \item[] Answer: \answerYes{}
    \item[] Justification: 
    The paper specifies the datasets, train/test settings, model variants, decoder settings, optimizer, learning-rate schedules, update counts, modality probabilities, and several hyperparameters in Section 4.1 and Appendix~\ref{app:expsetup}.

\item {\bf Experiment statistical significance}
    \item[] Question: Does the paper report error bars suitably and correctly defined or other appropriate information about the statistical significance of the experiments?
    \item[] Answer: \answerNo{}
    \item[] Justification: 
    The main benchmark tables report only word error rate (WER), following common practice in prior work. 
We do not report statistical significance or repeated-run variance because repeating these large-scale pretraining and finetuning experiments would be computationally very expensive.

\item {\bf Experiments compute resources}
    \item[] Question: For each experiment, does the paper provide sufficient information on the computer resources (type of compute workers, memory, time of execution) needed to reproduce the experiments?
    \item[] Answer: \answerYes{}
    \item[] Justification: The hardware used in our experiments and the corresponding training time are reported in Appendix~\ref{app:expsetup}.
    
\item {\bf Code of ethics}
    \item[] Question: Does the research conducted in the paper conform, in every respect, with the NeurIPS Code of Ethics \url{https://neurips.cc/public/EthicsGuidelines}?
    \item[] Answer: \answerYes{}
    \item[] Justification: We fully comply with the NeurIPS Code of Ethics.

\item {\bf Broader impacts}
    \item[] Question: Does the paper discuss both potential positive societal impacts and negative societal impacts of the work performed?
    \item[] Answer: \answerYes{} 
    \item[] Justification: 
    Broader impacts are discussed in Appendix~\ref{app:impact}.
    
\item {\bf Safeguards}
    \item[] Question: Does the paper describe safeguards that have been put in place for responsible release of data or models that have a high risk for misuse (e.g., pre-trained language models, image generators, or scraped datasets)?
    \item[] Answer: \answerNo{}
    \item[] Justification: 
    We do not consider this work to present unusually high misuse risks requiring dedicated technical safeguards. 
Nevertheless, any future release of code or models will be intended for legitimate research use, subject to applicable dataset licenses and relevant privacy considerations.

\item {\bf Licenses for existing assets}
    \item[] Question: Are the creators or original owners of assets (e.g., code, data, models), used in the paper, properly credited and are the license and terms of use explicitly mentioned and properly respected?
    \item[] Answer: \answerYes{}
    \item[] Justification: 
    We adhere to the usage restrictions of all assets and cite their sources and include their licenses
    in Appendix~\ref{app:expsetup}.

\item {\bf New assets}
    \item[] Question: Are new assets introduced in the paper well documented and is the documentation provided alongside the assets?
    \item[] Answer: \answerYes{}
    \item[] Justification: 
    Our released code and pretrained models are well documented, and the accompanying README files provide detailed instructions for reproducing our results.

\item {\bf Crowdsourcing and research with human subjects}
    \item[] Question: For crowdsourcing experiments and research with human subjects, does the paper include the full text of instructions given to participants and screenshots, if applicable, as well as details about compensation (if any)? 
    \item[] Answer:  \answerNA{} 
    \item[] Justification: We do not involve any crowdsourcing or research with human subjects.

\item {\bf Institutional review board (IRB) approvals or equivalent for research with human subjects}
    \item[] Question: Does the paper describe potential risks incurred by study participants, whether such risks were disclosed to the subjects, and whether Institutional Review Board (IRB) approvals (or an equivalent approval/review based on the requirements of your country or institution) were obtained?
    \item[] Answer: \answerNA{} 
    \item[] Justification: The paper does not involve new human-subject data collection, intervention, or participant studies by the authors.

\item {\bf Declaration of LLM usage}
    \item[] Question: Does the paper describe the usage of LLMs if it is an important, original, or non-standard component of the core methods in this research? Note that if the LLM is used only for writing, editing, or formatting purposes and does \emph{not} impact the core methodology, scientific rigor, or originality of the research, declaration is not required.
    \item[] Answer: \answerYes{} 
    \item[] Justification: LLM usage is part of the technical method. The paper includes an LLM-based decoder using Qwen-2.5-7B with an adaptor and LoRA, and this is described in both the main paper and Appendix.

\end{enumerate}

\end{document}